\documentclass[letterpaper]{article} 
\usepackage{aaai2026}  
\usepackage{times}  
\usepackage{helvet}  
\usepackage{courier}  
\usepackage[hyphens]{url}  
\usepackage{graphicx} 
\usepackage{natbib}  
\usepackage{caption} 
\usepackage{algorithm}

\usepackage{booktabs}
\usepackage{amsmath}
\usepackage{multirow}
\usepackage{pifont}
\usepackage{makecell}
\usepackage{algpseudocode, algorithmicx}
\usepackage{amssymb}
\newcommand{\cmark}{\ding{51}}  
\newcommand{\xmark}{\ding{55}}  

\usepackage{newfloat}
\usepackage{listings}
\DeclareCaptionStyle{ruled}{labelfont=normalfont,labelsep=colon,strut=off} 
\floatstyle{ruled}
\newfloat{listing}{tb}{lst}{}
\floatname{listing}{Listing}
\title{Vista: Scene-Aware Optimization for Streaming Video Question Answering Under Post-Hoc Queries}
\author{
     \textbf{Haocheng Lu$^{\rm 1, }$$^{\rm 2}$},
      \textbf{ Nan Zhang$^{\rm 2}$},
     \textbf{ Wei Tao$^{\rm 1, }$$^{\rm 2}$},
     \textbf{ Xiaoyang Qu$^{\rm 2*}$},
     \textbf{ Guokuan Li$^{\rm 1}$\thanks{ Corresponding authors.}},\\
     \textbf{ Jiguang Wan$^{\rm 1}$},
     \textbf{ Jianzong Wang$^{\rm 2}$}
}
\affiliations{

     \textsuperscript{\rm 1}Huazhong University of Science and Technology,
    \\
     \textsuperscript{\rm 2}Ping An Technology (Shenzhen) Co., Ltd.
     \\
     haocheng\_lu@hust.edu.cn, \{nzhang889, ericsullivan538, quxiaoy\}@gmail.com,
     \\
     \{liguokuan, jgwan\}@hust.edu.cn, jzwang@188.com
}

\usepackage{bibentry}

\begin{document}

\maketitle

\begin{abstract}

Streaming video question answering (Streaming Video QA) poses distinct challenges for multimodal large language models (MLLMs), as video frames arrive sequentially and user queries can be issued at arbitrary timepoints. Existing solutions relying on fixed-size memory or naive compression often suffer from context loss or memory overflow, limiting their effectiveness in long-form, real-time scenarios.We present Vista, a novel framework for scene-aware streaming video QA that enables efficient and scalable reasoning over continuous video streams. The innovation of Vista can be summarized in three aspects: (1) \textbf{Scene-aware segmentation. } Vista dynamically clusters incoming frames into temporally and visually coherent scene units. (2) \textbf{Scene-aware compression.} Each scene is compressed into a compact token representation and stored in GPU memory for efficient index-based retrieval, while the full-resolution frames are offloaded to CPU memory. (3) \textbf{Scene-aware recall.} Upon receiving a question, relevant scenes are selectively recalled and reintegrated into the model’s input space, enabling both efficiency and completeness. Vista is model-agnostic and integrates seamlessly with a variety of vision-language backbones, enabling long-context reasoning without compromising latency or memory efficiency. Extensive experiments on StreamingBench demonstrate that Vista achieves state-of-the-art performance, establishing a strong baseline for real-world streaming video understanding.

\end{abstract}


\section{Introduction}

Multimodal Large Language Models \cite{team2023gemini, team2024gemini, anthropic2024claude35, openai2024, zhang2025videollama, xu2025qwen2, tao2025madllm}  have achieved significant breakthroughs in a variety of fields, such as embodied AI, autonomous driving, and video-based dialogue systems. 

\begin{figure}
    \centering
    \includegraphics[width=\linewidth]{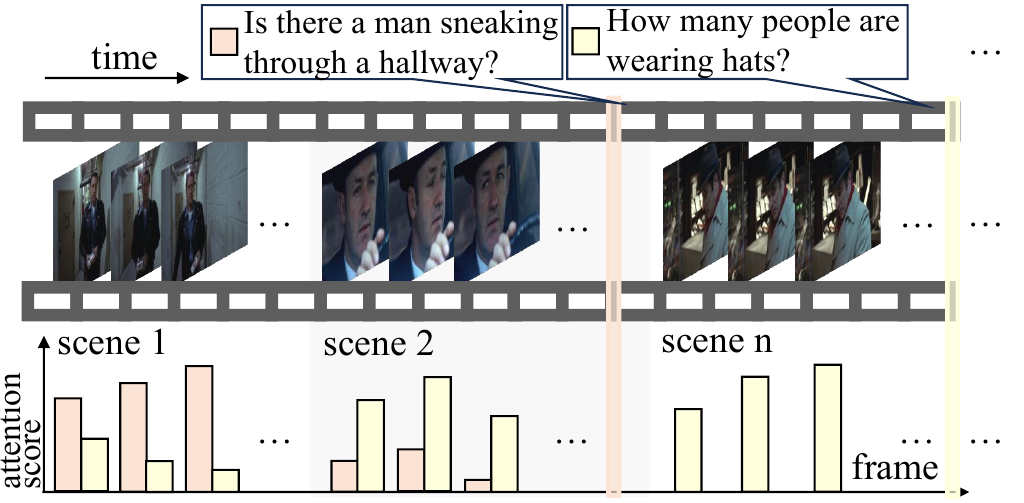}
    \caption{Illustration of the streaming video QA setting with post-hoc queries and scene-level attention. User questions may arrive at any time during video playback, and the model's attention distribution typically concentrates on a single coherent scene.}
    \label{fig:motivation}
\end{figure}

While progress in long-form video question answering has been impressive, most existing approaches are designed for offline inference, where the entire video and the corresponding question are simultaneously available for global analysis. However, this assumption breaks down in interactive real-time applications, where video streams arrive continuously and questions may be issued at arbitrary moments. In such streaming settings, the model must respond promptly without access to the full video context. This gives rise to two core challenges: (1) \textbf{Unbounded input length} --- Streaming videos are potentially infinite in duration, and sampling frames at a fixed rate can rapidly exhaust memory and computational resources; and (2) \textbf{Low-latency requirement} --- Timely responses are essential, which prohibits expensive full-sequence attention mechanisms at inference time. 

Consequently, streaming video QA requires fundamentally different system designs, ones that support continuous, query-agnostic encoding of incoming video streams, minimize memory usage over time, and allow for fast and effective retrieval upon question arrival. 

To overcome this limitation, we propose a novel framework that performs scene-level compression and indexing specifically designed for streaming video question answering. The core idea stems from the observation that, during inference, the model’s attention is typically concentrated on a limited number of semantically salient segments within the video, as illustrated in Figure~\ref{fig:motivation}. 

In summary, this paper makes the following contributions:

\begin{itemize}
    \item We propose a scene-aware segmentation and compression framework that organizes streaming video into temporally and semantically coherent scene units. Each scene is represented with a compact token-level summary, enabling effective compression without losing essential visual semantics.
    \item We introduce a scene-aware offloading and recall mechanism that stores high-resolution frames in CPU memory while retaining only compressed scene representations in GPU. At inference time, relevant scenes are selectively recalled based on the input query, allowing scalable long-context reasoning with minimal GPU memory overhead.
    \item Empirical results demonstrate the effectiveness of our approach in maintaining high accuracy while significantly reducing GPU memory usage and latency.
\end{itemize}
 Our method allows the model to perform lightweight attention over compressed scene tokens and retrieve full-resolution features on demand.

\section{Related Works}
Recent advances in vision-language models \cite{li2023blip, maaz2023video, liu2024llavanext, li2024llava, xu2024pllava} have substantially improved multimodal understanding, especially in Video QA. These models typically combine a pretrained visual encoder \cite{radford2021learning, zhai2023sigmoid} with a large language model through cross-modal alignment, enabling grounded responses to visual content. While effective for short clips, their direct application to long-duration or streaming videos is limited by inefficient retention and reasoning over temporally distant information.

\subsection{Long Video QA}
To address the limitations of vision-language models on extended temporal content, a growing body of research has focused on long video question answering, where the input spans hundreds or thousands of frames. In this setting, a central challenge lies in capturing long-range dependencies without incurring excessive memory and computation overhead. Several methods have been proposed to mitigate this, including token merging \cite{renggli2022learning, song2024moviechat+, li2024llama, weng2024longvlm, jin2024chat, xu2024slowfast, he2024ma}, sparse sampling \cite{nuthalapati2023coarse, tang2025adaptive}, KV pruning \cite{chen2024image} and memory retrieval \cite{buch2022revisiting, ram2023context, yu2023self}.

While these approaches improve efficiency in offline scenarios, they often assume full access to the video and the question during inference. This assumption is incompatible with real-time streaming settings. 

\subsection{Streaming Video QA}
Streaming Video Question Answering requires incremental processing of continuous video streams under strict memory and latency constraints, unlike offline methods that assume full access to both video and query. Existing approaches address this through different forms of compression, retrieval, and memory management. VideoStreaming \cite{qian2024streaming} introduces learnable modules for long-context modeling, while Flash-VStream \cite{zhang2024flash} and Video-LLaMA-Online \cite{chen2024videollm} compress incoming frames for scalable storage or high-frequency streaming. DisPider \cite{qian2025dispider} decouples perception and decision for real-time interaction. Retrieval-based methods offer a complementary direction: ReKV \cite{di2025streaming} retrieves KV-caches from CPU memory, LiveVLM \cite{ning2025livevlm} revisits prior segments via retrieval-augmented streaming, InfiniPot-V \cite{kim2025infinipot} compresses temporal caches, and StreamMem \cite{yang2025streammem} builds lightweight proxy memories. However, these methods largely operate at the frame or cache level and remain sensitive to temporal noise, leaving open challenges in achieving robust retrieval and efficient long-context reasoning.

\section{Preliminary}
\subsection{Streaming Video QA with Post-Hoc Queries}

\textbf{Streaming Video} refers to a continuous sequence of video frames $\{F_t\}_{t=1}^{T}$ arriving sequentially over time. At each timestep $t$, a frame $F_t$ is sampled and made available to the system at a fixed frame rate. Unlike offline video processing, where the entire clip is accessible in advance, streaming settings impose a strict temporal constraint where only current and past frames are observable, and future frames remain unknown during inference.

\noindent \textbf{Post-hoc Video Question Answering} is a variant of the video QA task where the query $Q$ arrives at an arbitrary time $t_q$ \emph{after} the video has already started streaming. Formally, given a query $Q$ arriving at $t_q$, the model must provide an answer based solely on the sequence of frames observed up to that point, i.e., $\{F_t\}_{t=1}^{t_q}$. 

The post-hoc setting introduces a unique set of challenges. In traditional sparse attention or key-frame selection methods, frame importance is often determined by the query, allowing the model to retain only query-relevant frames. However, in the post-hoc scenario, the query $Q$ is unavailable during frame observation. Consequently, the model cannot estimate frame importance in advance and must operate in a \emph{query-agnostic} fashion during streaming.

Furthermore, due to the limited GPU memory capacity, it is infeasible to retain all frames $\{F_t\}$ in high resolution throughout the entire streaming process. This prevents the model from deferring frame selection until query arrival. Instead, it must balance three competing demands:
\begin{itemize}
    \item \textbf{Relevance:} Retain sufficient visual content that could be relevant to future unknown queries;
    \item \textbf{Efficiency:} Discard or compress frames early to manage memory usage;
    \item \textbf{Responsiveness:} Upon query arrival, retrieve or reconstruct the most informative visual evidence with minimal latency.
\end{itemize}

These constraints define the core difficulty of streaming video QA under post-hoc queries: the model must make decisions under temporal uncertainty, memory pressure, and strict latency constraints, all while ensuring semantic fidelity to support accurate question answering.

\begin{figure*}
    \centering
    \includegraphics[width=\linewidth]{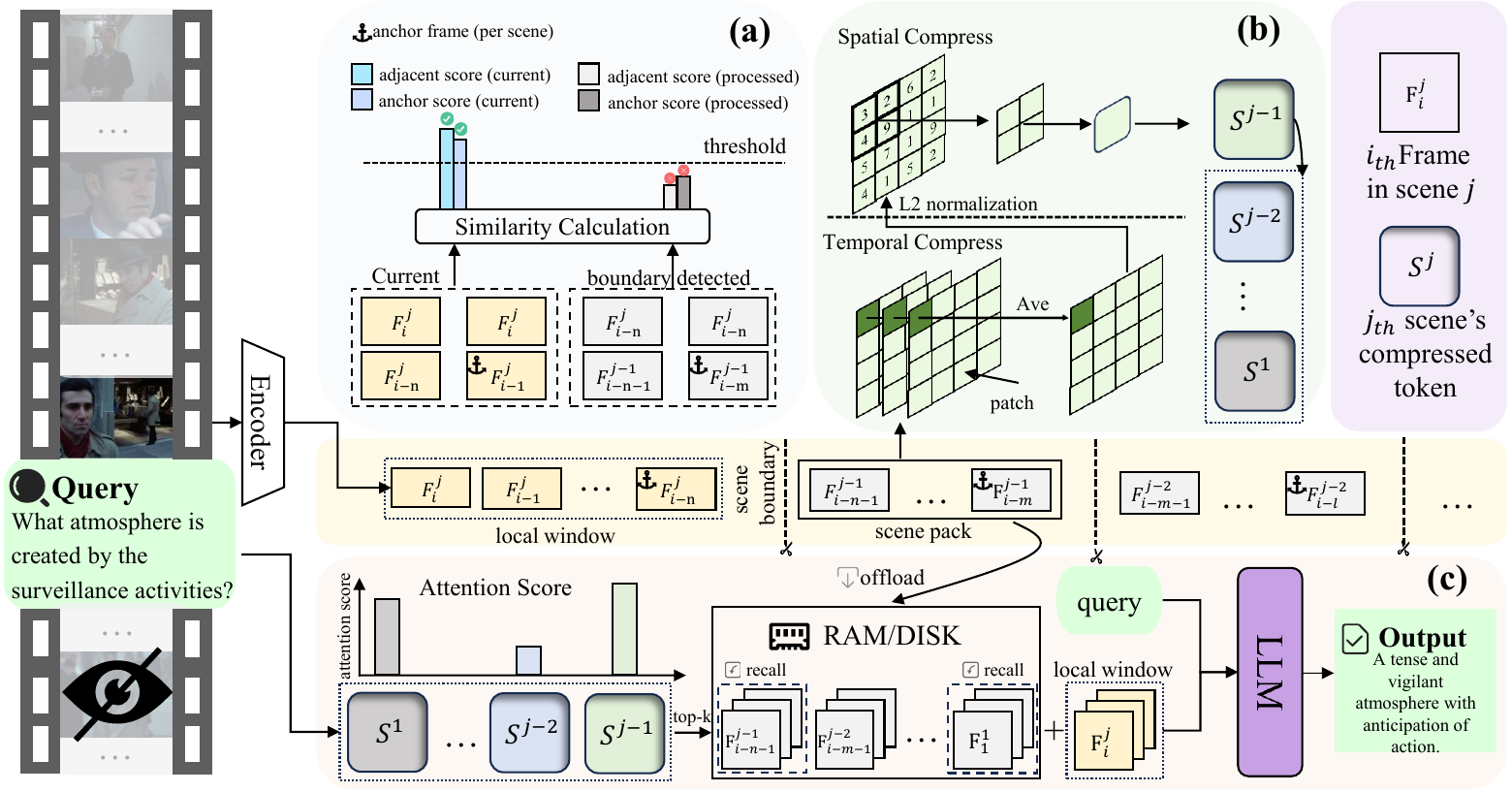}
    \caption{The framework of Vista. (a) Scene-aware segmentation. (b) Scene-aware compression. (c) Scene-aware recall.}
    \label{fig:framework}
\end{figure*}

\section{Method}
\subsection{Overall Framework}
Figure \ref{fig:framework} illustrates the overall framework of Vista. Unlike traditional video question answering settings where the entire video and the question are both known in advance, our task operates under a streaming paradigm: video frames arrive sequentially at a fixed frame rate, and the query is revealed only at an arbitrary time point during inference. This constraint prevents preemptive content-aware frame selection based on the question, making efficient memory management a critical challenge.

To address this, Vista employs a scene-aware segmentation and compression framework that groups streaming video frames into temporally and semantically coherent scenes. Each incoming frame is encoded and appended to a local sliding window, with scene boundaries detected based on frame-level similarity metrics. Upon identifying a scene boundary, all frames within the preceding scene are compactly summarized into a token-level representation, typically via pooled embeddings, enabling effective compression without losing essential visual semantics. Meanwhile, the original high-resolution frame features are offloaded to CPU memory, maintaining a minimal GPU memory footprint.

Once the query is issued, we perform attention-based scene retrieval by computing relevance scores between the query representation and the compressed scene tokens. The top-ranked scene tokens are then restored by fetching their corresponding full-resolution features from CPU or disk storage. These are concatenated with the current contents of the local window to form the final multimodal video input sequence.

This sequence is subsequently fed into the vision-language model to generate the output. By controlling the number of full-resolution frames, Vista effectively maintains a bounded GPU memory usage while retaining the ability to dynamically recover relevant information on demand.

\subsection{Scene-aware Segmentation}

In streaming video scenarios, efficient and accurate detection of scene boundaries is crucial for effective compression and memory management. Since the query is unknown during streaming, key frame selection cannot rely on semantic relevance to the question. Therefore, an online, unsupervised approach based on visual similarity patterns is adopted to identify scene transitions.

Let $F_a$ denote a fixed anchor frame representing the start of the current scene, and $F_i$ the latest incoming frame in the video stream. 

A scene boundary is declared only if both similarity scores fall below a predefined threshold $\tau$:

\begin{equation}
    \mathcal{B}(F_i) = \mathbb{I}\left[ 
        \underbrace{\mathcal{S}_{\text{anchor}}(F_i) < \tau}_{\text{Anchor condition}} 
        \land 
        \underbrace{\mathcal{S}_{\text{adj}}(F_i) < \tau}_{\text{Adjacency condition}}
    \right]
\end{equation}
, where $S_{\text{anchor}}(F_i)$ denotes the similarity between frame $F_i$ and the corresponding anchor frame within the current scene, and $S_{\text{adj}}(F_i)$ measures the similarity between frame $F_i$ and its immediately preceding frame. $\tau$ is a predefined threshold controlling the boundary sensitivity, and $\mathbb{I}[\cdot]$ is the indicator function that returns 1 if the condition holds and 0 otherwise.

To ensure smooth transitions between scenes and to accommodate gradual visual changes, a temporal overlap between consecutive scenes is introduced. This overlap allows a small buffer of frames to be shared by adjacent scenes, mitigating abrupt boundary effects and preserving temporal coherence during downstream processing.

\subsection{Scene-aware Compression}

Before the question is provided, the system continuously encodes the incoming video stream frame-by-frame. Let $F_i$ denote the $i$-th frame in the raw video sequence. During streaming processing, each frame is additionally assigned to a detected scene, denoted as $F_i^j$, indicating that frame $F_i$ belongs to the $j$-th scene segment. This notation reflects the segmentation of the continuous video stream into consecutive scene units.

To exploit the temporal locality of question-relevant frames, which tend to cluster within scenes
, a \textit{local window} buffer is maintained to store recent frames for scene grouping. Specifically, when a new frame $F_i$ arrives, the current local window is defined as: $ \mathcal{L} =  [F_{i-n}^j, \cdots, F_{i-1}^j] $
where all frames in the window share the same scene index $j$.

The decision to perform compression is triggered upon detecting a scene boundary between frames $F_{i-1}$ and $F_i$. In such a case, the local window containing frames of scene $j$ is finalized: all frames within this window are aggregated as a scene unit and offloaded to RAM for storage, while the new frame $F_i$ starts a fresh local window for the next scene. Concurrently, a compact \textit{scene token} is generated to summarize the information of the completed scene, facilitating efficient future retrieval based on query relevance.

To accurately and efficiently encode each scene, a \textit{Temporal-Spatial Compression} strategy is proposed, leveraging redundancy along both temporal and spatial dimensions. This strategy can be summarized as three steps:

\begin{itemize}
    \item \textbf{Temporal Compression:} Given the set of frames $\{F_i^j\}_{i=1}^m$ within the $j$-th scene, average pooling is applied along the temporal axis for each spatial patch independently. Then the results of all the spatial patches are integrated to achieve the temporally compressed feature map $F_{temp}$. This step exploits the high correlation of adjacent frames within a scene, effectively reducing temporal redundancy.
    
    \item \textbf{Spatial Compression:} The temporally compressed feature map $F_{temp}$ is reshaped into a two-dimensional grid of spatial tokens. To ensure compatibility with patch sizes, zero-padding is applied if necessary. To emphasize salient regions, we introduce a weighting scheme based on L2 norms: each spatial patch's L2 norm serves as an importance measure, guiding weighted averaging within a sliding spatial window.

    \item \textbf{Final Aggregation:} Following spatial weighting and fusion, a second average pooling operation aggregates the spatial tokens into a single compact vector, producing the final scene token that encapsulates both global semantic context and local discriminative details.
\end{itemize}

\begin{algorithm}[t]
\caption{Scene-aware Compression}
\label{alg:scene_token_generation}
\begin{algorithmic}[1]
\Require Set of frames $\{F_i\}_{i=1}^m$ within one scene, height of sliding window $h$, width of sliding window $w$
\Ensure Compressed scene token $T$ representing the scene

\For{each spatial patch location $(x, y)$}
    \State $F[x,y] \leftarrow \frac{1}{m} \sum_{i=1}^{m} F_i[x,y]$ 
\EndFor
\State Integrate all the patches $F_{temp} \leftarrow \{F[x,y]\}$ 
\State Reshape $F_{\text{temp}}$ into 2D grid of spatial tokens $S$
\State $H, W \leftarrow \text{height, width of } S$

\For{$s = 1$ to $H$} 
    \For{$t = 1$ to $W$} 
        \If{$s+h > H$ or $t+w > W$}
            \State apply zero padding
        \EndIf
        \State $\mathcal{W}_{s,t} \leftarrow \{S_{i',j'}\}_{i'=s,j'=t}^{s+h,t+w}$ 
        \State $w_{s,t} \leftarrow \|\mathcal{W}_{s,t}\|_2$
        \State $S_{\text{fused}}^{(s,t)} \leftarrow \frac{\sum_{i'=s,j'=t}^{s+h,t+w} w_{i',j'} \cdot S_{i',j'}}{\sum_{i'=s,j'=t}^{s+h,t+w} w_{i',j'}}$
    \EndFor
\EndFor

\State $T \leftarrow \frac{1}{H\times W} \sum_{s=1,t=1}^{H,W} S_{\text{fused}}^{(s,t)}$ 

\State \Return $T$
\end{algorithmic}
\end{algorithm}

This hierarchical compression pipeline significantly reduces memory footprint while preserving critical information for downstream query-driven retrieval. Algorithm \ref{alg:scene_token_generation} outlines the core steps of the temporal-spatial compression process.

\begin{table*}[t]
\centering
\begin{tabular}{l|c|ccccc|ccccc}
\toprule
& \textbf{RT} & \multicolumn{5}{c|}{\textbf{Omni-Source Understanding}} & \multicolumn{5}{c}{\textbf{Contextual Understanding}}  \\
\textbf{Model} & all & ER & SCU & SD & MA & all & ACU & MCU & SQA & PO & all\\ 
\midrule
Human & 91.46 & 88.00 & 88.24 & 93.60 & 90.27 & 90.26 & 88.80 & 90.40 & 95.00 & 100 & 93.55 \\
\midrule
\multicolumn{10}{l}{\textbf{Proprietary MLLMs}} \\ 
Gemini 1.5 pro & 75.69 & 46.80 & 39.60 & 74.90 & 80.00 & 60.22 & 51.41 & 40.73 & 54.80 & 45.10 & 48.73 \\
GPT-4o & 73.28 & 41.20 & 37.20 & 43.60 & 56.00 & 44.50 & 41.20 & 38.40 & 32.80 & 56.86 & 38.70 \\
Claude 3.5 Sonnet 20 & 72.44 & 31.60 & 34.00 & 32.80 & 48.80 & 36.80 & 38.40 & 34.80 & 34.40 & 64.71 & 37.70 \\
\midrule
\multicolumn{10}{l}{\textbf{Streaming MLLMs}} \\ 
Flash-VStream & 23.23 & 25.91 & 24.90 & 25.60 & 28.40 & 26.00 & 24.80 & 25.20 & 26.80 & 1.96 & 24.12 \\
VideoLLM-online & 35.99 & 31.20 & 26.51 & 24.10 & 32.00 & 28.45 & 24.19 & 29.20 & 30.80 & 3.92 & 26.55 \\
Dispider & 67.63 & 35.46 & 25.26 & 38.57 & 43.34 & 35.66 & 39.62 & 27.65 & \textbf{34.80} & 25.34 & 33.61 \\
\midrule
LLaVA-OneVision-7B & {70.92} & 40.00 & {24.80} & {31.20} & {44.40} & {35.10} & {32.40} & {35.60} & {30.80} & {\textbf{33.20}} & {33.00}  \\
\textbf{+Vista(ours)} &  
\makecell{\textbf{71.36} \\ \scriptsize{(\textbf{+0.44})}}& 
\makecell{\textbf{46.40} \\ \scriptsize{(\textbf{+6.40})}} & \makecell{\textbf{37.20} \\ \scriptsize{(\textbf{+12.40})}} & \makecell{\textbf{43.60} \\ \scriptsize{(\textbf{+12.40})}} & \makecell{\textbf{74.00} \\ \scriptsize{(\textbf{+29.60})}} & 
\makecell{\textbf{50.30} \\ \scriptsize{(\textbf{+15.20})}}&
\makecell{\textbf{43.20} \\ \scriptsize{(\textbf{+10.80})}} & \makecell{\textbf{36.80} \\ \scriptsize{(\textbf{+1.20})}} & 
\makecell{34.40 \\ \scriptsize{(+3.60)}} & 
\makecell{29.60 \\ \scriptsize{(-3.60)}} &
\makecell{\textbf{36.00} \\ \scriptsize{(\textbf{+3.00})}}
\\
\midrule
Video-LLaMA2-7B & 52.6 & {38.18} & {23.47} & {34.09} & {39.04} & {32.93} & {26.90} & {27.61} & {27.60} & {0.00} & {20.53} \\
\textbf{+Vista(ours)} &  
\makecell{52.92 \\ \scriptsize{(+0.32)}}& 
\makecell{39.60 \\ \scriptsize{(+1.42)}} & 
\makecell{26.80 \\ \scriptsize{(+3.33)}} & 
\makecell{33.60 \\ \scriptsize{(-0.49)}} & 
\makecell{45.60 \\ \scriptsize{(+6.56)}} & 
\makecell{36.40 \\ \scriptsize{(+3.47)}}&
\makecell{28.00 \\ \scriptsize{(+1.10)}} & 
\makecell{32.00 \\ \scriptsize{(+4.39)}} & 
\makecell{32.00 \\ \scriptsize{(+4.40)}} & 
\makecell{\hspace{0.25em}0.00\hspace{0.25em} \\ \scriptsize{(+0.00)}} & 
\makecell{23.00 \\ \scriptsize{(+2.47)}} \\
\bottomrule
\end{tabular}
\caption{Comparison of video-language models on StreamingBench across various capabilities. Most of the data is cited from StreamingBench \cite{lin2024streamingbench}.}
\label{tab:streamingbench}
\end{table*}

\subsection{Scene-aware Recall}
Efficient retrieval of relevant visual content from memory is crucial at query time $t_q$. Given the constrained GPU memory during inference, especially for storing full-resolution video frames, our goal is to selectively restore only the most relevant scenes while adhering to strict memory usage bounds.

Let $Q$ denote the incoming textual query, and let $\mathcal{S} = \{T_1, T_2, \ldots, T_N\}$ be the set of previously compressed scene tokens, where each $T_j$ is a temporally and spatially condensed representation of scene $j$ whose frames have been offloaded to CPU memory or disk.

The query is first embedded using a language encoder $\psi(\cdot)$, yielding the query embedding: $\mathbf{q} = \psi(Q)$.

To assess the relevance between the query and each scene, a similarity function, typically implemented as scaled dot-product attention, is employed to compute attention scores. The attention score $\alpha_i$ is computed as $\alpha_i = \mathbf{q} T_i^\top$.

The top-$k$ most relevant scene tokens are then selected according to their attention scores. Let $\mathcal{I}_k$ denote the indices of the selected top-$k$ tokens:
\begin{equation}
    \mathcal{I}_k = \left\{ j \mid \alpha_j \in \text{TopK}(\{\alpha_i\}_{i=1}^N, k) \right\}
    \label{eq:topk_selection}
\end{equation}

For each selected index $j \in \mathcal{I}_k$, its corresponding full-resolution frames $\mathcal{F}_j = \{F_1^j, \ldots, F_m^j\}$ are retrieved from disk or CPU memory.

The final multimodal input to the vision-language model is constructed by aggregating the following components:
\begin{itemize}
    \item The union of full-resolution frames from the top-$k$ retrieved scenes: $\bigcup_{j \in \mathcal{I}_k} \mathcal{F}_j$;
    \item The most recent uncompressed frames in the local processing window, denoted as: $\mathcal{L}$;
    \item The tokenized query $Q$.
\end{itemize}

Let $\mathcal{V}_{\text{final}} = \left( \bigcup_{j \in \mathcal{I}_k} \mathcal{F}_j \right) \cup \mathcal{L}$ denote the video input sequence composed of the elements in recalled frames and local window. The overall input to the vision-language model is then defined as: $ \text{Input}_{\text{VLM}} = (\mathcal{V}_{\text{final}}, Q) $.

This dynamic retrieval mechanism, which integrates both query-relevant full-resolution scenes and the most recent local frames, enables the model to maintain high inference accuracy while remaining within strict GPU memory constraints.

\section{Experiments}

\subsection{Experimental Setup}
\subsubsection{Benchmarks}
To evaluate the effectiveness of our method on the task of streaming video question answering, we conduct experiments on both online and offline video benchmarks, including \textbf{StreamingBench}~\cite{lin2024streamingbench}, \textbf{EgoSchema}~\cite{mangalam2023egoschema}, and \textbf{MLVU}~\cite{zhou2024mlvu}.

StreamingBench is a comprehensive benchmark suite designed to assess the diverse capabilities of streaming video understanding models. 

MLVU serves as a multi-task benchmark for evaluating the long-range comprehension abilities of video-language models across various levels of semantic understanding in extended video content.

EgoSchema is a large-scale diagnostic benchmark focusing on temporal event comprehension and schema-based reasoning in egocentric videos, offering fine-grained evaluations for long-form video understanding.

\subsubsection{Hardware}

We conduct our experiments on four NVIDIA 4090D GPUs, each containing 24GB of memory, along with an i9-14900K CPU and 125GB of RAM. All experiments use greedy decoding with temperature set to 0, making the generation deterministic.

\subsection{Performance on StreamingBench} 

We conducted a thorough evaluation of our method on the StreamingBench benchmark. The results, presented in Table~\ref{tab:streamingbench}, demonstrate that our approach achieves highly competitive and, in several key areas, state-of-the-art performance compared to existing streaming MLLMs. Our framework significantly enhances the capabilities of the base models (e.g., LLaVA-OneVision-7B), showcasing its effectiveness and general applicability.

\begin{figure}
    \centering
    \includegraphics[width=\linewidth]{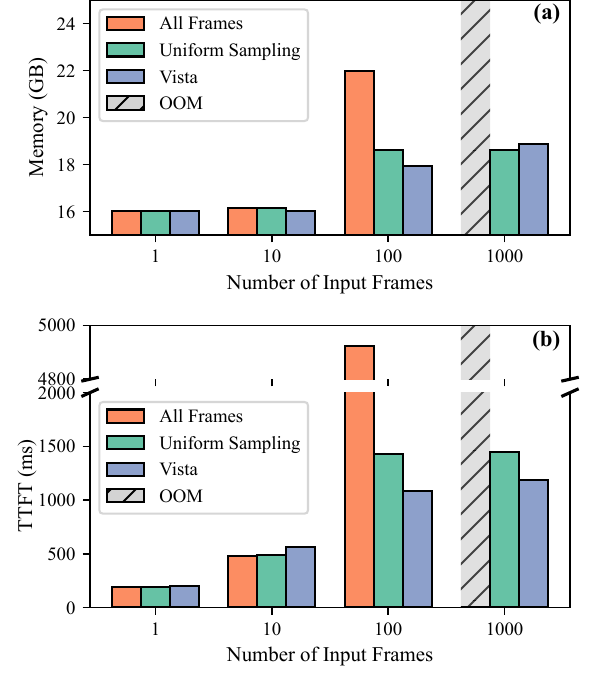}
    \caption{Performance comparison with increasing input frames across different strategies: (a) memory usage; (b) inference latency.}
    \label{fig:efficiency}
\end{figure}

Real-Time Visual Understanding (RT) assesses a model’s capability to perceive and reason over visual content up to a query timestamp, emphasizing immediacy via temporal cues. Our method significantly improves RT performance, achieving 71.36\% (+0.44\%) when combined with LLaVA-OneVision, surpassing other streaming baselines.


Vista demonstrates its most significant advantages in Omni-Source Understanding. The most notable result appears in the Multimodal Alignment (MA) task, where Vista achieves 74.00\%, a +29.60\% absolute improvement over the baseline. This score not only surpasses all other streaming models but also exceeds proprietary models such as GPT-4o (56.00\%). We also observe clear gains in Emotion Recognition (46.40\%, +6.40\%) and in SCU/SD (37.20\% and 43.60\%), highlighting Vista’s ability to maintain precise temporal and semantic alignment across modalities.


For Contextual Understanding tasks requiring reasoning over continuous video streams, Vista also shows clear gains. It improves ACU performance to 43.20\% (+10.80\%), surpassing streaming MLLMs such as Dispider (39.62\%), and achieves consistent improvements in MCU and SQA. These benefits mainly stem from the staged recall mechanism, which efficiently retrieves relevant historical context for each query.

To evaluate the scalability and efficiency of our method, we compared it with two baselines: uniform sampling input and full frame input under varying numbers of input frames. As shown in Figure~\ref{fig:efficiency}, our method consistently maintains low memory usage and stable latency (TTFT) even as the number of input frames increases. 

Overall, our framework significantly advances streaming MLLMs. The consistent low latency and memory footprint under varying input sizes further demonstrate its practical scalability. Notably, the state-of-the-art performance in multimodal alignment and other key tasks validates the effectiveness and broad applicability of our approach in streaming video understanding.

\subsection{Performance on Offline Video Benchmarks} 

\begin{figure}
    \centering
    \includegraphics[width=0.98\linewidth]{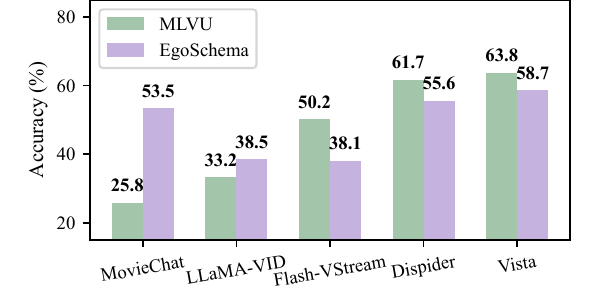}
    \caption{Offline video question-answering on different long-form benchmarks. Most of the data is cited from Dispider \cite{qian2025dispider}.}
    \label{fig:Offline}
\end{figure}

We evaluate vista on two representative long-video QA benchmarks: \textbf{MLVU} and \textbf{EgoSchema}, and compare it with both long-video understanding models and streaming-oriented models.

As shown in Figure~\ref{fig:Offline}, vista achieves the notable accuracy on both benchmarks, reaching 63.8\% on MLVU and 58.7\% on EgoSchema. Compared to other strong offline MLLMs designed for long video QA, vista yields substantial improvements, highlighting its superior ability to understand long-range temporal contexts.

Vista also surpasses leading streaming models such as Dispider and Flash-VStream, demonstrating that the proposed scene-aware mechanism not only enables real-time interaction but also generalizes well to offline long-form scenarios. These results validate the robustness and versatility of our temporal modeling framework across different video QA settings.

\begin{table}[H]
\centering
\begin{tabular}{lccc|c}
\toprule
\textbf{Base} & \textbf{Seg.} & \textbf{Comp.} & \textbf{Recall} & \textbf{Accuracy (\%)} \\
\midrule
 \cmark    & \xmark & \xmark & \xmark & 40.00 \\
 \cmark    & \xmark & \cmark & \cmark & 38.80 \\
 \cmark    & \cmark & \xmark & \cmark & 42.00 \\
 \cmark    & \cmark & \cmark & \xmark & 44.00 \\
 \cmark & \cmark & \cmark & \cmark & \textbf{46.40} \\
\bottomrule
\end{tabular}
\caption{Ablation study of scene-aware segmentation, scene-aware compression and scene-aware recall strategies.}
\label{tab:ablation}
\end{table}

\subsection{Ablation Study}





To evaluate the contribution of each component, we conduct an ablation study on the \textbf{Emotion Recognition (ER)} task, examining the effects of \emph{scene-aware segmentation}, \emph{scene-aware compression}, and \emph{scene-aware recall} (Table~\ref{tab:ablation}).

The base model using uniformly sampled frames achieves 40.00\% accuracy. Applying compression and recall without segmentation slightly decreases performance to 38.80\%, indicating that compression without semantic grouping may introduce misaligned cues.

Using segmentation alone improves accuracy to 42.00\%, showing that temporal-semantic partitioning provides a beneficial structural prior. Adding compression to segmented scenes further raises performance to 44.00\%, suggesting that compact representations within coherent scenes effectively preserve salient information.

Combining all three modules achieves the best result (46.40\%), demonstrating their complementary roles and the overall effectiveness of the proposed framework.

\begin{figure}
    \centering
    \includegraphics[width=\linewidth]{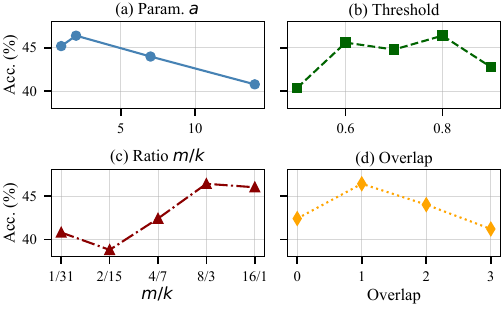}
    \caption{Sensitivity analysis of four different hyperparameters for streaming video QA: (a) Window Size $a$; (b) Similarity threshold  $\tau$; (c) Compression-recovery ratio $m / k$; (d) Overlap step size.}
    \label{fig:hyperpara}
\end{figure}

To evaluate the robustness of our framework, we perform a hyperparameter sensitivity analysis on four key variables: the spatial compression window size $a$, the segmentation similarity threshold $\tau$,the scene capacity–recall limit pair $m/k$, and the temporal overlap step size, as shown in Figure~\ref{fig:hyperpara}. Here, $a$ controls how many spatial patches are aggregated within each frame during compression, $\tau$ determines when a new scene is started based on frame-to-anchor similarity, the pair $m/k$ specifies the maximum number of frames allowed in each scene $m$ and the maximum number of scenes that can be recalled during question answering $k$, and the overlap step size governs how many frames are shared between consecutive scenes.

For the spatial window size $a$ (Figure~\ref{fig:hyperpara}(a)), increasing $a$ from 1 to 2 improves performance (45.2\% to 46.4\%) by enabling broader within-frame aggregation, while overly large windows (e.g., $a=7$) oversmooth spatial details, indicating that moderate values work best.

For the segmentation threshold $\tau$ (Figure~\ref{fig:hyperpara}(b)), low thresholds (0.5) cause under-segmentation and high thresholds (0.9) over-fragment scenes. The optimal value of 0.8 strikes a balance between coherence and compactness.

For the scene capacity–recall pair $m/k$ (Figure~\ref{fig:hyperpara}(c)), we vary $m$ and $k$ under a fixed total capacity. Very small $m$ (e.g., 1) produces fragmented context, while large $m$ (e.g., 16) merges overly diverse content. An intermediate configuration (e.g., $m=8, k=3$) achieves the best performance (46.4\%).

Temporal overlap (Figure~\ref{fig:hyperpara}(d)) further smooths scene transitions. A small overlap (step=1) improves accuracy (42.4\% to 46.4\%), whereas larger overlaps add redundancy without additional gains.

Overall, the analyses demonstrate that both scene-aware segmentation and controlled scene capacity are crucial for robust performance, and Vista remains stable across diverse hyperparameter settings.





\begin{figure}
    \centering
    \includegraphics[width=\linewidth]{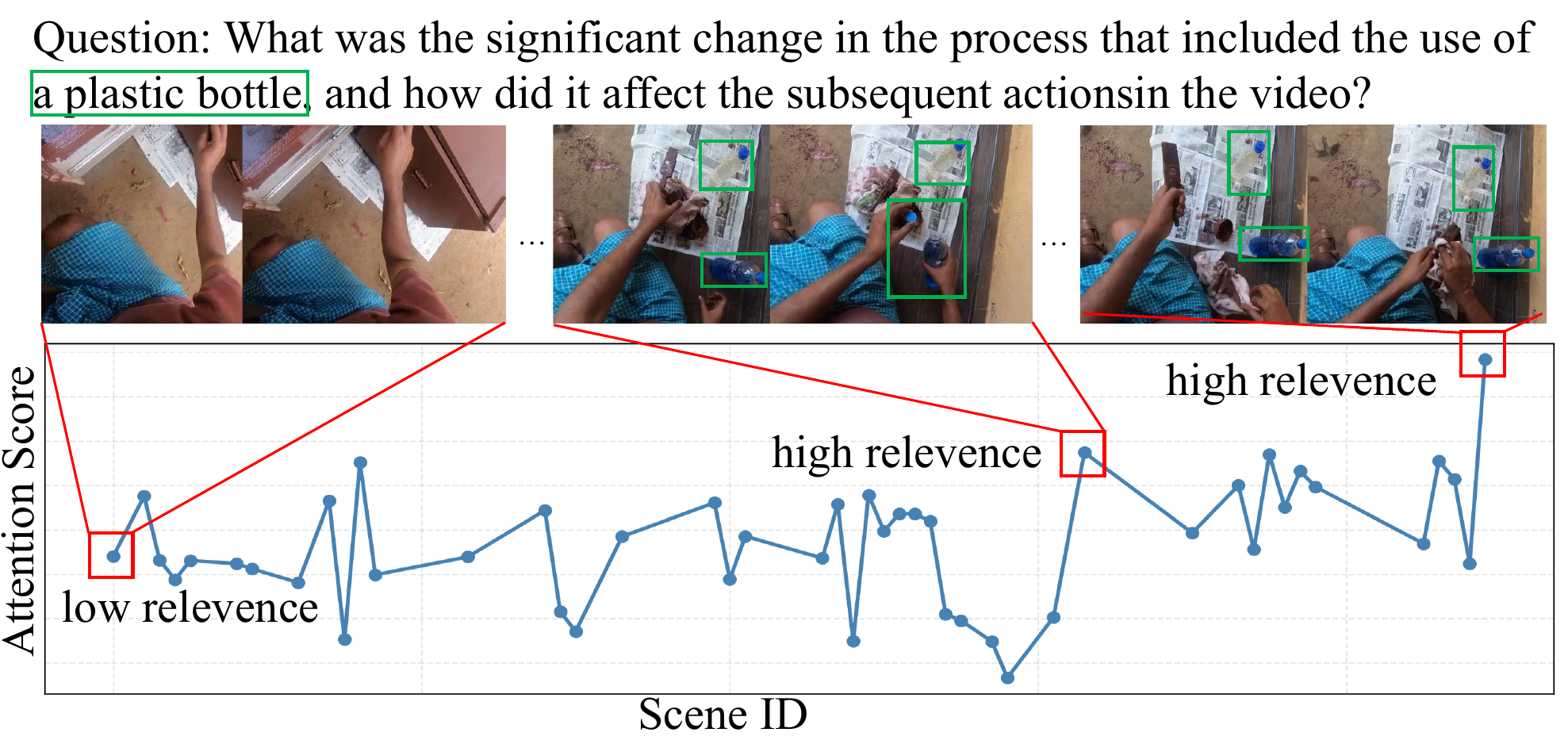}
    \caption{Visualization of the scene-aware recall strategy when the query arrives. Scenes with higher relevance scores are selectively recalled and often contain critical information necessary to answer the question.}
    \label{fig:visualiaztion}
\end{figure}


\subsection{Visualization of Scene Recall}

To further illustrate the effectiveness of our scene-aware recall strategy, we visualize the recall scores across different scene segments for a representative video in Figure~\ref{fig:visualiaztion}. As shown, our model assigns higher relevance scores to scenes that are semantically aligned with the given question, such as the use of a plastic bottle. These high-scoring segments are subsequently selected and aggregated for downstream reasoning. The qualitative alignment between the recalled scenes and the query demonstrates the model's ability to selectively retain question-relevant visual content, enabling more accurate and context-aware video question answering. This visualization confirms that our recall mechanism successfully filters out irrelevant or redundant scenes while emphasizing those that are temporally and semantically critical for answering the question.

While Vista effectively recalls semantically relevant scenes, certain limitations remain.
In highly dynamic scenes with rapid motion or abrupt transitions, accurate boundary detection becomes difficult, and Vista temporarily falls back to single-frame recall.
Conversely, in long static scenes, a maximum scene length is enforced to avoid memory overflow.
In complex scenarios with overlapping events, recall accuracy may degrade due to mixed scene representations.

\section{Conclusion}
This paper proposes a novel real-time Streaming Video QA framework based on scene-aware optimization. Our method introduces three key innovations: scene-aware segmentation that dynamically groups incoming frames into coherent scenes based on visual-temporal patterns, scene-aware compression that converts these scenes into compact GPU-stored tokens while offloading raw frames to CPU, and scene-aware recall that selectively retrieves and reintegrates relevant scenes when answering queries. These innovations achieves both computational efficiency and information completeness. Experiments on standard streaming video QA benchmarks confirm that our approach outperforms state-of-the-art methods.

\section*{Acknowledgements}
This work was sponsored by the Shenzhen-Hong Kong Joint Funding Project (Category A) under Grant No. SGDX20240115103359001, the National Key Research and Development Program of China under Grant No.2023YFB4502701, Shandong Provincial Natural Science Foundation under Grant No. ZR2024LZH004.

\bibliography{aaai2026}

\end{document}